%% file: paper.tex
\documentclass{sig-alternate-05-2015}

\usepackage{balance}
\usepackage{color}
\usepackage{graphicx}
\usepackage{mediabb}
\usepackage{subfigure}
\usepackage{amsmath,amssymb,bm}
\usepackage{times}
\usepackage{url}
\usepackage{nameref}
\usepackage{here}
\usepackage{comment} 
\usepackage{algorithm}
\usepackage{algorithmic}

\begin{document}


\doi{}

\isbn{}



%

\title{Deep Recurrent Neural Network for\\ Mobile Human Activity Recognition
with High Throughput
}


%
%
%
%

\numberofauthors{3} 
%
\author{
%
%
\alignauthor
Masaya Inoue\titlenote{He is the corresponding author, and executed theory construction, simulation, and verification of the results.}\\
       \affaddr{Kyushu Institute of Technology}\\
       \affaddr{1-1 Sensui ,Tobata}\\
       \affaddr{Kitakyushu ,Fukuoka ,Japan}\\
       \email{q344203m@mail.kyutech.jp  }
\alignauthor
Sozo Inoue
\\
       \affaddr{Kyushu Institute of Technology}\\
       \affaddr{1-1 Sensui ,Tobata}\\
       \affaddr{Kitakyushu ,Fukuoka ,Japan}\\
       \email{~~~sozo@mns.kyutech.ac.jp}
\alignauthor Takeshi Nishida
\\
       \affaddr{Kyushu Institute of Technology}\\
       \affaddr{1-1 Sensui ,Tobata}\\
       \affaddr{Kitakyushu ,Fukuoka ,Japan}\\
       \email{~nishida@cntl.kyutech.ac.jp}
}

\date{30 July 1999}


\maketitle
\begin{abstract}
\input{abst.tex}
\end{abstract}


\input{intro.tex}
\input{related.tex}
\input{principle.tex}
\input{eval.tex}
\input{result.tex}

\input{discussion.tex}
\input{conclusion.tex}

\section{Acknowledgments}
This work was supported by JSPS KAKENHI Grant Number 26280041.


\end{document}

%% file: abst.tex


In this paper, we propose a method of human activity recognition with high throughput from
raw accelerometer data applying a deep recurrent neural network (DRNN),
and investigate various architectures and its combination to find the
best parameter values. The ``high throughput'' refers to short time at a time of recognition.
We investigated various parameters and architectures of the DRNN by using the training dataset of 432 trials with 6 activity classes from 7 people.
The maximum recognition rate was 95.42\% and 83.43\% against the test
data of 108 segmented trials each of which has single activity class and
18 multiple sequential trials, respectively. Here, the maximum
recognition rates by traditional methods were 71.65\% and 54.97\% for
each. In addition, the efficiency of the found parameters was evaluated by using additional dataset.
Further, as for throughput of the recognition per unit time, the constructed DRNN was requiring only 1.347 [ms], while the best traditional method required 11.031 [ms] which includes 11.027 [ms] for feature calculation.
These advantages are caused by the compact and small architecture of the constructed real time oriented DRNN.

\thispagestyle{empty}

%% file: intro.tex

\section{Introduction}
\label{sec:Introduction}
The recognition of human activity is a task that is applicable to
various domains, such as health care, preventive medicine, and elderly
care. In addition, with the rapid spread of devices with built-in
sensors such as smartphones recently, the cost of sensing devices has
fallen significantly. As a result, researches on mobile activity
recognition have been actively
conducted \cite{avci2010activity}.

In traditional activity recognition schemes, researchers have frequently used a machine learning method, such as decision tree, $k$-nearest neighborhood, naive Bayes, support vector machine, and random forest, to recognize activities from a feature vector extracted from signals in a time window by using statistic values or Fourier transformation.

Recurrent neural networks (RNN) is the name of neural networks that include a directed closed cycle. The RNN is suitable for handling time-series data, such as audio and video signals, and natural language. In recent years, the hierarchical multi-layered convolutional neural network (CNN) has achieved noteworthy results in areas such as image processing, and is drawing attention to the method called deep learning. In this trend, because the RNN also has a deep layer for temporal direction, it has come to be captured as a deep learning method.

Compared to traditional activity recognition methods which are input feature vectors, in deep learning, the original data can be directly input. This allows the calculation of feature vectors to be skipped at the time of training and recognition, so that a speed-up can be expected, especially in the recognition.
At the same time, we can also expect the recognition result to the to be highly accurate by virtue of the deep learning.


In this paper, we propose a method of human activity recognition from raw accelerometer data applying a RNN, and investigate various architectures and its combination to find the best parameter values. 

By using a human activity sensing consortium (HASC) open dataset, the
recognition ability of the constructed RNN was evaluated. We used the
training dataset of 432 segmented trials with 6 activity classes from 7
people, and it was confirmed that the maximum recognition rate was
95.42\% against the test data of 108 segmented trials each of which has
single activity class. While the recognition rate of traditional method
was 71.65\%. Moreover, the maximum recognition rate was 83.43\% against
the test data of 18 multiple sequential trials, and while the
recognition rate of the traditional method was 54.97\%. Where ``a
trial'' means one sequential data sample such as a segmented data or a
sequence data. 
Moreover, a network reconstructed with the parameters investigated by
using HASC dataset by using the human activity recognition (HAR) open
dataset, 95.03\% recognition rate was achieved.
\\~~
Further, for the throughput of the recognition per unit time, the proposed method was fast requiring only 1.347 [ms],
while the existing method required 11.031 [ms] which includes 11.027
[ms] for feature extraction. 
Notice that this calculation time achieved by only using CPU. The fast response advantage is caused by the number of weights less than 10 \% of the traditional method \cite{ordonez2016deep}.

The contribution of this study includes the following three points:

\begin{enumerate}
\item 
      In order to construction of a fast response classifier oriented
      real time execution, we adopted a RNN architecture and evaluated its advantages compared with traditional methods.
\item To improve the accuracy of the RNN, various parameters were
      explored to investigate the factors that affect the
      accuracy. We used the two types of dataset.
\item The throughput for recognition with the RNN was evaluated, and it was shown to be faster than the existing method which includes feature calculation.

\end{enumerate}

%% file: related.tex
\section{Background and Related Work}\label{sec:related}

Recently, many studies on the mobile activity recognition are carried out
\cite{Bulling2014,Lane2010a}.
For activity recognition technology, techniques for various applications, such as sports
\cite{Strohrmann2011,kunze2006towards},
skills assessment \cite{ladha2013climbax}, 
detection and evaluation of walking \cite{mazilu2015wearable},
medical analyses
and nursing activity analysis
\cite{Inouye2001}
were proposed.
In these techniques, a machine learning method to recognize the activity, such as decision tree, $k$-nearest neighborhood, naive Bayes, support vector machine, and random forest, is often employed as a basic technique, after the feature vector have been extracted from the signals by statistics or Fourier transformation by taking a time window \cite{Bao2004a}.

Since activity recognition handles sequential data, techniques for sequential data, such as the hidden Markov model (HMM)
\cite{Kim2010}
 and conditional random fields
 \cite{Zhan2014},
 which are used in speech recognition and natural language processing,
 were proposed. In addition, while studies on completing the required activity recognition in real time have been conducted in a few work
\cite{Krishnan2014},
 these studies focused mainly on how to reduce the feature calculation by shifting the feature vectors.
Similarly, many studies have been focused on reducing the resources required for executing the feature processing, e.g. 
\cite{Saeedi2014,bhattacharya2014using}
 .


Moreover, in recent years, several activity recognition methods which
use deep CNN have been proposed, and they have been confirmed that they
can achieve high accuracy recognition than the traditional methods
\cite{yang2015deep}. However these method require the time window to generate certain length segmentation of time series signal. Moreover, in general, CNNs have huge number of connection between inner layers. These features of CNNs are not suitable for real time execution of mobile devices. 

An RNN can be used as a learning method and an estimator, and is
 suitable for handling time-series data, such as audio and video
 signals, and natural language.
Early RNNs included the fully recurrent network developed in 1980, an interconnected type network such as the Hopfield network announced by John J. Hopfield in 1982,
and so on. Then, hierarchical RNNs, such as the Elman network and the Jordan network were developed 
in the early 1990s \cite{NN}. The Elman network has a state feedback,
and the Jordan network has a recurrent connection for the output
feedback.
This feedback contributes in order to extract features of dynamics of input signal. The RNN executes calculation processing of large network that led to the time direction at the training phase, and executes fast sequential calculation processing at the recognition phase.

Recently, many of the methods using CNNs and RNNs aims to recognize by using raw signal directly without extraction of the feature vectors in advance.



In the method using a combination of CNN and RNN \cite{ordonez2016deep}, the network
achieves further high accuracy recognition by the feature extraction ability of dynamics in the RNN. On the other hand, the adoption of the CNN architecture causes the increasing of recognition rate, the increasing of the computational cost, and the utilization of the time window.

The RNN is a high throughput network architecture that can deal with raw sensor data without feature extraction and can recognize by thorough fast sequential processing. In this paper, we propose a method to execute training and recognition of the RNN (i.e. deep RNN) which has multi internal layer by using raw acceleration data without feature extraction aiming at a high-precision activity recognition with high throughput.

%% file: principle.tex
\section{Recurrent Neural Network}
\label{sec:rnn}
In the following, the basic processing methods for execution of training and recognition of a RNN are explained.

\subsection{Deep recurrent neural model}
\label{sec:forward}
\begin{figure}[b]
 \centering
\includegraphics[width=32mm]{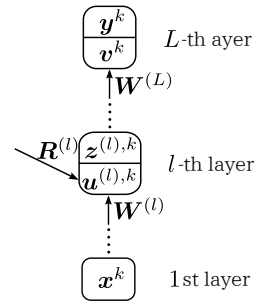}
 \caption{Schematic representation of the DRNN}
 \label{fig:RNNの略図}
\end{figure}

Let us assume a deep RNN (DRNN) with $L$ layers, as shown in Fig.~\ref{fig:RNNの略図}. This
network is an Elman-type network in which internal layers are
completely connected at the same hierarchy in the time direction. Here,
${\bm u}^{(l),k} = [~u^{(l),k}_1~ u^{(l),k}_2~ \cdots u^{(l),k}_j \cdots
u^{(l),k}_J~]^T $ is the input vector of the $l$-th layer at time $k$
and ${\bm z}^{(l),k} = [~z^{(l),k}_1~ z^{(l),k}_2~\\ \cdots z^{(l),k}_j~ \cdots
z^{(l),k}_J~]^T $ is the output vector of the $l$-th layer at time
$k$. A pair of each elements of the input and output vectors is called a
unit. $j$ is an arbitrary unit number of the $l$-th layer and $J$ is
the total number of units.
We assume ${\bm x}^k= {\bm z}^{(1),k}$ in the
input layer, and ${\bm v}^k={\bm u}^{(L),k}$ and ${\bm y}^k={\bm z}^{(L),k}$ in the output layer.
In addition, in the following,
 the arbitrary unit numbers (and the total numbers of units) of the $(l-1)$-th layer are represented by $i$ (and $I$, respectively).
 At this time, the input propagation weight from the $(l-1)$-th layer to
 the $l$-th layer is represented by ${\bm W^{(l)}}(\in \mathbb{R}^{J
 \times I})$ and ${\bm R}^{(l)}(\in \mathbb{R}^{J \times J})$ is the recurrent weight in the $l$-th layer $(l=2,\cdots,L-1)$,
where $j'$ is an arbitrary unit number of the $l$-th layer before one time unit.
At this time, the components of ${\bm u}^{(l),k}$ are given by

\begin{equation}
\label{eq:eq1}
 u^{(l),k}_j = \sum_{i}^I w^{(l)}_{ji}z^{(l-1),k}_i + \sum_{j'}^J r^{(l)}_{jj'}z^{(l),k-1}_{j'}.
\end{equation}
Here $w^{(l)}_{ji}$ and $r^{(l)}_{jj'}$ represent the element of ${\bm
W^{(l)}}$ and ${\bm R}^{(l)}$, respectively. 
The elements of the output vector of the $l$-th layer are expressed as
\[
 z^{(l),k}_j = f^{(l)}(u^{(l),k}_j),
\]
where $f^{(l)}(\cdot)$ is called the \emph{activation function}, and functions such as the sigmoid function $f(u) = {\rm tanh}(u)$, logistic sigmoid function $f(u) = 1/(1+e^{-u})$, and rectified linear unit (ReLU) function $f(u) = {\rm max}(u,0)$ are frequently used.

Here, for simplicity, by introducing the 0-th weight $w^{(l)}_{j0}$ and the 0-th unit $z^{(l-1),k}_0=1$, biases can be collectively described as
\begin{equation}
\label{eq:eq5}
{\bm  z}^{(l),k} = {\bm f}^{(l)}({\bm W}^{(l)}{\bm z}^{(l-1),k} + {\bm R}^{(l)}{\bm z}^{(l),k-1}),
\end{equation}
where ${\bm f}({\bm a}) = [~f(a_1)~f(a_2) \cdots f(a_N)~]^T$.
From this equation, it is possible to obtain the output of an arbitrary time by shifting $k$.
However, since elements of ${\bm v}^k$
has no recurrent connection, it is the same as the first term of Formula~(\ref{eq:eq1}).
Therefore, the final output vector ${\bm y}$ is derived by
\begin{equation}
\label{eq:eq6}
{\bm  y}^k = {\bm f}^{(L)}({\bm v}^k)= {\bm f}^{(L)}({\bm W}^{(L)}{\bm z}^{(L-1),k}).
\end{equation}


\subsection{Learning method}
\label{sec:Learning method}


\subsubsection{Error function}
\label{sec:誤差関数}
When performing multi-class classification into class
$C_1, \cdots , C_h,\\ \cdots , C_H$, by using the softmax function, let the output of
the $h$-th unit of the output layer be the following equation.
Further, the output $y_h$ of the individual unit means a probability belonging to class $C_h$.
\begin{equation}
 \label{eq:eq7}
y_h\equiv z^{(L)}_h= \frac{\exp (u_h)}{\sum _{q=1}^H \exp (u_q)}=p(C_h|{\bm x}).
\end{equation}
When an input ${\bm x}$ is given, this probability $y_h$ is classified
into the largest class, and
\begin{equation}
 \label{eq:eq9}
E({\bm w})= -\sum_{n=1}^N \sum_{h=1}^H d_{nh} \log y_h({\bm x}_n;{\bm w})
\end{equation}
is defined as the error function, and updating the variable vector ${\bm
w}$ to minimize this becomes the learning policy. The ${\bm d_n}$
represents $n$-th supervised vector and the $d_{nh}$ represents $h$-th
elements of dn. The value of element is set as 1 if the $h$-th element
corresponds to the class of ${\bm d_n}$, and otherwise it is set as 0. Eqn. (\ref{eq:eq9}) is called the cross entropy function.

\subsubsection{Mini-batch stochastic gradient descent}
\label{sec:ミニバッチ確率的勾配降下法}
It is possible to use the gradient descent method to minimize the error function. Let $D$ be the number of elements of ${\bm w}$; the gradient of the error function is expressed by
\[
 \nabla E\equiv \frac{\partial E({\bm w})}{\partial {\bm w}}
 =\biggl[\frac{\partial E({\bm w})}{\partial w_1} \cdots \frac{\partial E({\bm w})}{\partial w_D}\biggr]^T.
\]
The gradient descent method searches the local minimum value in the neighborhood
by repeating many times to change ${\bm w}$ in the negative gradient direction by a very small amount. Let the weight in the $t$-th time of repeat be ${\bm w}^t$; then, it becomes
\begin{equation}
 \label{eq:eq10}
{\bm w}^{t+1}={\bm w}^t-\epsilon \nabla E,
\end{equation}
where $\epsilon$ is a parameter called the learning rate.

The mini-batch stochastic gradient descent method collects a small number of sample sets $B^t$ (called a \emph{mini-batch}) in each repeat to calculate the gradient using the average
\begin{equation}
 \label{eq:eq11}
E^t({\bm w})=\frac{1}{||B^t||}\sum_{n\in B^t} E_n({\bm w})
\end{equation}
of error for each sample $n$ among them.
It is known that the local solution avoidance performance is high because the calculation converges quickly with this method.

As the learning rate $\epsilon$ is high, learning becomes faster;
however, if it is too high, since it vibrates near the local minimum
value of the error function, an adjustment method of the learning rate,
called Adaptive moment estimation (Adam), is employed~\cite{adam}.



\subsubsection{Back Propagation Through Time (BPTT)}
 \label{sec:BPTT}
\begin{figure}[b]
 \centering
 \includegraphics[width=72mm]{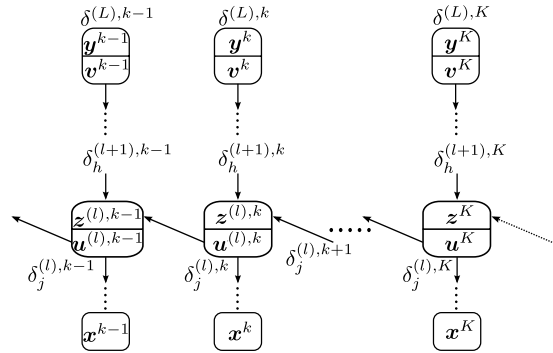}
 \caption{Back propagation in DRNN}
 \label{fig:RNNでの逆伝播}
\end{figure}

The back propagation through time (BPTT) method can be applied the error back-propagation method to the developed DRNN regarded as a large NN expanded in the time direction.
The technique is described below. First, we introduce here a quantity called \emph{delta} for the unit $j$ of layer $l$
\begin{equation}
 \label{eq:eq12}
\delta^{(l),k}_j \equiv \frac{\partial E_n({\bm w})}{\partial u^{(l),k}_j},
\end{equation}
This value can also be derived from the values of the $(l+1)$-th layer of the same time and the time $k+1$ of the same layer. Let the derivative of the function $f(u)$ be $f'(u)$; then \begin{equation}
 \label{eq:eq17}
\delta^{(l),k}_j = \Biggl( \sum_h w^{(l+1)}_{hj} \delta^{(l+1),k}_h + \sum_{j'}
r^{(l)}_{j'j} \delta^{(l),k+1}_{j'} \Biggr) f'(u^{(l),k}_j)
\end{equation}
holds.
The gradient can be calculated based on these equations. The weights $w^{(l)}_{ji}, r^{(l)}_{jj'}$ to be updated become
\begin{equation}
 \label{eq:eq18}
\frac{\partial E({\bm w})}{\partial w^{(l)}_{ji}}=\sum_{k=1}^K \frac{\partial
E({\bm w})}{\partial u^{(l),k}_j}\frac{\partial u^{(l),k}_j}{\partial w^{(l)}_{ji}} =\sum_{k=1}^K \delta^{(l),k}_j z^{(l-1),k}_i,
\end{equation}

\begin{equation}
 \label{eq:eq19}
\frac{\partial E({\bm w})}{\partial r^{(l)}_{jj'}} = \sum_{k=1}^K \frac{\partial
E({\bm w})}{\partial u^{(l),k}_j}\frac{\partial u^{(l),k}_j}{\partial r^{(l)}_{jj'}} =\sum_{k=1}^K \delta^{(l),k}_j z^{(l),k-1}_j.
\end{equation}
Using these relations,
the gradients can be calculate by propagating the deltas inversely from output to input, as shown in Fig.~\ref{fig:RNNでの逆伝播}.


Here, the calculation amount becomes too enormous for practical use in the normal BPTT to calculate the gradient by dating back all the times. Therefore, the truncated BPTT method \cite{T-BPTT}, which sets the time to date back to an appropriate constant to perform BPTT for each the time, is used.

\subsection{Long short-term memory (LSTM)}
\label{sec:LSTM}
 Long short-term memory (LSTM) is a type of NN model for time series
 data.
 It is utilized mainly to replace some units of the RNN to solve the problems of
an input/output weight conflict~\cite{LSTM97} which is the conflicts between the input from the previous layer and the recurrent value,
  and vanishing/exploding gradient problem~\cite{clipping} where a delta vanishes or explodes by the deep backward propagation.
In this section, we describe the structure of LSTM which solves input/output weight conflicts and vanishing gradients, and gradient clipping method to avoid exploding gradients. 

\subsubsection{Structure of LSTM}
\label{LSTMの構造}

\begin{figure}[b]
 \centering
 \includegraphics[width=68mm]{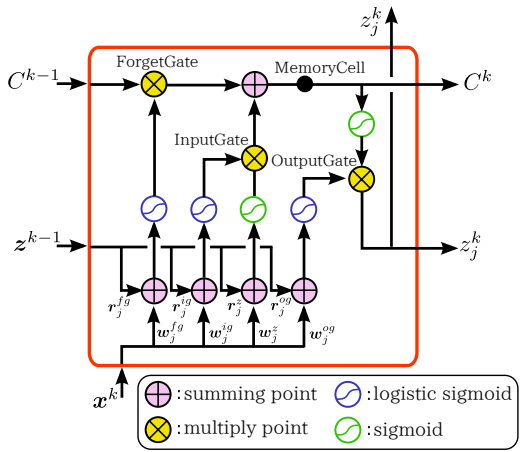}
 \caption{Structural drawing of LSTM}
 \label{fig:LSTMの構造図}
\end{figure}

A structural diagram of the LSTM is shown in Fig. \ref{fig:LSTMの構造図}. A structure for storing the internal state, called a memory cell, is provided in the LSTM, allowing it to perform the controls, such as whether to write the information to the cell, read the information from the cell, or delete the information of the cell.

The structure horizontally propagating straight in the top portion of Fig. \ref{fig:LSTMの構造図} is for solving the vanishing gradient problem and is called the Constant Error Carousel (CEC). It is possible to solve the vanishing gradient problem by introducing the CEC when back-propagating the error in the recurrent direction \cite{LSTM97}. In addition, with this structure, the state $C$ inside the internal layer unit is transmitted at the next time. Although a memory cell is also clearly shown in Fig. \ref{fig:LSTMの構造図}, in fact the state is preserved through the entire structure of the CEC.

The input gate and output gate are for eliminating input and output
weight conflicts \cite{LSTM97}. Let us consider the input gate as an
example. First, the output vector ${\bm z^{k-1}}(\in \mathbb{R}^{J
\times 1})$ before 1 time and the input vector
${\bm x^{k}}(\in \mathbb{R}^{I \times 1})$ of the present time
multiplied by the transmission weight ${\bm r^{ig}_j}(\in \mathbb{R}^{1
\times J}), {\bm w^{ig}_j}(\in \mathbb{R}^{1 \times I})$ are summed to
pass through the logistic function. This is expressed by the following
equation, where the logistic function is expressed by $\sigma$ and the
input gate bias by $b^{ig}$:
\begin{equation}
 \label{eq:eq21}
\phi^k=\sigma({\bm w^{ig}_j}{\bm x^{k}}+{\bm r^{ig}_j}{\bm z^{k-1}}+b^{ig}).
\end{equation}
Because the logistic function returns a value in the range of 0 to 1, when multiplied by the original transmission input at the next input gate, it controls “how much of the input to pass”. When $\phi^k$ is 0, it does not pass the input completely, and it passes all the inputs when it is 1. The same operation is performed in the output gate. By providing a gate that performs such an operation, it is possible to determine whether to memorize a state or to read the memorized state in accordance with the input value or the output value to the internal layer unit.

The purpose of the forget gate is to determine
whether or not to forget the memorized state \cite{LSTM99}. However, the memory forgetting mentioned here
does not refer to inheriting the value of the memory cell before one time. The mechanism of the forget gate is the same as that of the input and output gates, as described above. The
forget gate operates, for example, to perform efficient learning even in cases, such as that where the pattern of the time-series data is changed suddenly to a pattern having no correlation with the previous context.

By introducing the LSTM, the weights to be updated will increase. More specifically, the bias weight in addition to $r^{ig}, w^{ig}$ for the transmission to the input gate, $r^{og}, w^{og}$ for the transmission to the output gate, and $r^{fg}, w^{fg}$ for the transmission to the forget gate will increase
by each gate weight. These weights can also be updated by transmitting the delta inside the LSTM block using the back-propagation method.

\subsubsection{Gradient clipping}
\label{sec:Gradient clipping}
The exploding gradient problem is solved with a technique called gradient clipping.

Gradient clipping is a method of correcting the $L2$ norm of the gradient so that it does not exceed the threshold value \cite{clipping}. Specifically, when letting the threshold value be $c$,
\begin{equation}
 \label{eq:eq22}
 \|\nabla E\| \geq c
\end{equation}
is met, and a new value is assigned to the gradient as
\begin{equation}
 \label{eq:eq23}
\nabla E \leftarrow \frac{c}{\|\nabla E\|} \nabla E.
\end{equation}

\subsection{Avoiding overfitting}
\label{sec:過学習による汎化性能の低下}
Learning, as in NNs, by adopting the error function as an optimization function may cause overfitting, in which a model that captures too much peculiarities of the training data and does not fit to the new test data is generated.
As methods to avoid overfitting, regularization and dropout are available, as described in the following.

\subsubsection{Regularization}
\label{sec:正則化}
Overfitting is liable to occur when the degree of freedom of the network is too high for the training data. However, in many cases, the training data cannot be easily increased, and the degree of freedom of the network should not be easily reduced, because it is deeply involved in the expressive power of recognition. Therefore, a regularization method to mitigate overfitting by providing some type of constraint on the learning parameter is required. The gradient clipping described in Section \ref{sec:Gradient clipping} is an exploding gradient problem-solving technique as well as a regularization technique.

\subsubsection{Dropout}
\label{sec:ドロップアウト}
Dropout, which was developed recently, is a widely used overfitting avoidance technique. At the time of training, the units of the internal and output layers are disabled by selecting them at a constant rate $p$. That is, learning is performed as if they did not exist from the beginning. On this occasion, selection of the unit to be disabled is performed at every time to update the weight. At the recognition, all the units are used to perform the forward propagation calculation. However, the outputs of the units belonging to the target of disabled layers are uniformly multiplied by $p$ at the time of training \cite{dropout}.\\

As mentioned previously, there are many parameters in the RNN; even if
 we just use the RNN, 
 trial and error for setting the parameters depending on the problem will be required. Therefore, in the task of activity recognition, an examination of the parameters, such as the numbers of layers and units, truncated time, and dropout rate, should contribute to the study of the activity recognition using the RNN.

%% file: eval.tex

\section{Activity Recognition using RNN}
\label{sec:method}

We applied the DRNN described in Section \ref{sec:rnn} to human activity recognition to verify its accuracy and performance.

The items to be verified were are as follows:
\begin{enumerate}
\item Does the recognition accuracy increase as compared to that of other algorithms?
\item Is any influence exerted on the accuracy when some parameters are changed?
\item How long is the throughput time of the recognition as compared with other algorithms?

\end{enumerate}

\subsection{Dataset}

\label{sec:HASCコーパスと問題設定}
The HASC corpus is a dataset for machine learning gathered and
distributed by HASC \cite{kawaguchi2011hasc}, distributed
at a state with a detailed label attached to the data measured by
sensors mounted on a mobile device. In this study, we used a part of the
acceleration signals of the HASC corpus as a dataset. 
The dataset are divided into ``segmented data'' and ``sequence data'', the former includes single activity in one trial and the latter includes multiple consecutive activities.
The details of these two types of data are shown
in Table \ref{tab:使用したHASCデータセットの詳細}.
The segmented data are suitable for use as training data because they are able to label easily. On the other hand, since the sequence data are constructed by seamless measurement of human activities, these are resemble actual human activities.

\begin{table}[b]
  \caption{Details of HASC dataset}
  \label{tab:使用したHASCデータセットの詳細}
  \centering
 \fontsize{8pt}{10pt}\selectfont
  \begin{tabular}{c|c|c} \hline
&Segmented data&Sequence data \\ \hline
\parbox[t]{0.2\columnwidth}{Signal in one
   \\measurement}&\parbox[t]{0.3\columnwidth}{time~[{\rm s}], X~axis~[G],
       \\Y~axis~[G], Z~axis~[G]}&\parbox[t]{0.3\columnwidth}{time~[{\rm s}], X~axis~[G],\\Y~axis~[G], Z~axis~[G]} \\ \hline

\parbox[t]{0.20\columnwidth}{Frequency}&100 [Hz]&100 [Hz] \\ \hline
Targeted activity&\parbox[t]{0.3\columnwidth}{``stay", ``walk", ``jog"\\
       ``skip" ``stair~up",\\``stair down"}\vspace{0.015cm}&\parbox[t]{0.3\columnwidth}{``stay",``walk",``jog"\\
       ``skip" ``stair~up",\\``stair down"} \\ \hline
Measurement time&20~[{\rm s}]&120~[{\rm s}] \\ \hline
\parbox[t]{0.3\columnwidth}{Type of Activity \\in one measurement}\vspace{0.015cm}&1&6 \\ \hline
Number of person&7&7 \\ \hline
Number of trials&540&18 \\ \hline
{\color{black}Type}&Single
       activity&Multiple activity\\ \hline

  \end{tabular}
\end{table}

\subsubsection{Cross validation}

In the evaluation, we divided the segmented data into the training data
of 432 trials and the test data of 108 trials so that the number of
samples in each activity class balances each other.

Based on the data, we evaluated three types of  accuracy:
\begin{description}
\item[Training accuracy] Perform training with the training data, and recognize with the training data.
\item[Test accuracy] Perform training with the training data, and recognize with the segmented test data.
\item[Sequence accuracy] Perform training with the training data, and recognize with the sequence data.
\end{description}
Note that, because of the design of HASC dataset, for both sequence data and test data, the same person could be included.

As a measure of accuracy, the proportion of samples successfully recognized in the evaluation samples was used.

\subsubsection{Additional dataset}

As an additional dataset to examine the generality of our method, we
adopted the Human Activity Recognition using
Smartphones dataset (\emph{HAR dataset})\cite{anguita2013public} in the UCI Machine Learning Repository, and applied the best parameters found in the HASC dataset.
The sensor data were collected using smartphones equipped with a three-axis accelerometer and a gyroscope.
The smartphones were attached on the waists of the 30 persons.
They have six types of activity classes, which are ``Standing", ``Sitting", ``Laying", ``Walking", ``Walking downstairs", and ``Walking upstairs", and compiled as sequential data.
For cross validation, we used the first three fourth samples as training data, and last one fourth as test data.

\subsection{DRNN-based activity recognition}

\label{sec:構築したRNNモデル}

In order to perform high throughput activity recognition for each time
by using the three-axis acceleration of a smartphone as the direct
input, we constructed a DRNN such that the three-axis acceleration data
of each time corresponded to the three-dimensional input layer, and six
activity classes to the six-dimensional output layer. Each unit of each intrnal layer was an LSTM unit. The activation function of the output layer and the error function were defined by a softmax function and a cross entropy function, respectively. The truncated BPTT under the mini-batch stochastic gradient descent method was used to update the weights at the time of training. The number of internal layers, the number of units inside the internal layer, the number of times dating back was performed by truncated BPTT (called truncated time), the maximum gradient $c$, and also the dropout probability $p$ were set to be variable in order to search the most appropriate value experimentally. The details of this DRNN are summarized in Table \ref{tab:RNNの詳細}.

This network outputs an activity class, which corresponds to an element
having the largest value among the elements of the output vector
obtained when an input vector is input, as the recognized result.

\begin{table}[b]
\centering
\fontsize{8pt}{10pt}\selectfont
\caption{Details of DRNN}
\label{tab:RNNの詳細}
\begin{tabular}{c||c}
Setting items&Detail\\ \hline
Activation function of output layer& Softmax \\
Error function       &Cross entropy \\
Type of internal layer unit &LSTM            \\
Mini-batch size    & 20                \\
No. of time stamps in a mini-batch  &$K'$=1200\\
Initial Weights & random $[-0.1,0.1)$　\\
Initial bias & None \\
Learning rate adjustment     &Adam\\ 
Input dimension     &3 \\
Output dimension     &6 \\
\end{tabular}
\end{table}

\begin{figure}[t]
 \centering
 \includegraphics[width=55mm]{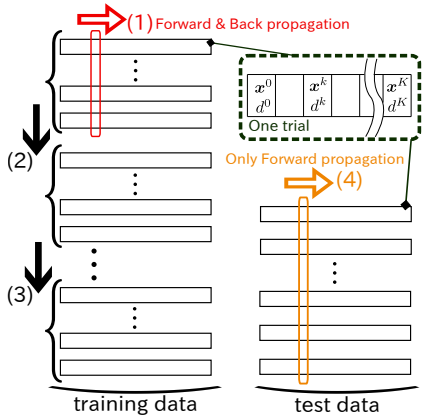}
 \caption{Flow of an epoch}
 \label{fig:1epochの流れ}
\end{figure}

A flow of the process of training and evaluation will be described below. The outline is also shown in Fig. \ref{fig:1epochの流れ}.

\begin{description}
\item[(0)] Shuffle all the trials of training data and divide them into mini-batch sets of 20 trials.
\item[(1)] For the first mini-batch,
\begin{enumerate}
\item Take the time $k$ at random.
\item Let the truncated time be $T$, and divide the time range for truncated BPTT into $[k,k+T-1], [k+T, k+2T-1], \cdots, [k+K'-T,  k+K']$, where the final value of the range was set as $K'=1200$.
\item For each range, obtain an error function from the input and output to update the weights by performing the error back-propagation.
\end{enumerate}
\item[(2)] Perform the same processing as (1) for the subsequent mini-batch.
\item[(3)] Call a period until the processing for all mini-batches is complete \emph{1 epoch}.
\item[(4)] Obtain the accuracy for the test data to valuate the generalization performance.
\end{description}

Repeat (1) to (4), and stop the training after repeating a sufficient
number of epochs. training phase ends in up here. After the training phase, the activity recognition of the sequence data are executed by using only the forward propagation of the learned model. Record the accuracies of training data, test data, and
sequence data in each epoch to plot the changes.


Chainer \cite{tokui2015chainer}, provided by Preferred Networks, Inc., was used to implement the
DRNN. Chainer is a framework (FW) for NN. In Chainer, various NN models can be flexibly written in the Python
language. The experiment was conducted in an environment as shown in
Table \ref{tab:計算機環境}.\\~~
Here, the training time was reduced by parallel processing using the GPU, and  the CPU was used for evaluating of the throughput of the constructed DRNN. This strategy is based on a following policy; the RNNs is trained by a large-scale computer architecture, and its execution is done by a standard type mobile terminal.

\begin{table}[b]
\centering
 \fontsize{8pt}{10pt}\selectfont
\caption{Computing environment}
\label{tab:計算機環境}
\begin{tabular}{c||c} \hline
OS&Ubuntu14.04LTS (64-bit) \\ \hline
CPU&Intel Corei5-4590 3.3 GHz \\ \hline
RAM    &DDR3-1600 24 GB \\ \hline
GPU     &NVIDIA Quadro K2200   \\ \hline
FW     &Chainer 1.5.1 \\ \hline
Python Ver.  &2.7.6       \\ \hline
CUDA Ver.  &7.0       \\ \hline
\end{tabular}
\end{table}

\subsection{Comparative methods}
As comparative methods, decision tree, support vector machine (SVM), and random forest were used.

The comparative methods, rather than the raw sensor data, and require time windows to calculate the feature vectors.
Therefore, we extracted feature vectors from the three-axis accelerometer data. For the sensor data, time windows of 5 s were extracted, shifting every 2.5 [s], as in Bao et al. \cite{Bao2004a}.

For each time window, we calculated 27 feature values, following the studies in \cite{Zhang2012,Zhang2011, ubicomp15sozo}.
The number of feature variables used in each time window was 27, including: (1-3) mean value of each axis, (4-6) variance of each axis, (7) mean sum of the absolute values of each axis, (8-9) first and second eigenvalue of the covariance matrix between the axes, (10) sum of the vertical component ratios for the intensity, (11-13) covariance ratio in the $x$ and $y$-direction for the $z$-component variance of each axis, (14-16) variance ratio of the back and forth difference in the $x$ and $y$-direction for the variance of the back and forth difference in the $z$-direction of each axis, (17-19) mean FFT-domain energy of each axis, (20) mean FFT-domain energy of the intensity, (21-23) FFT-domain entropy of each axis, (24) FFT-domain entropy of the intensity, (25) number of mean crosses of the mean intensity, (26) number of crosses of the zone of the mean intensity $\pm 0.1$[G], and (27) number of samples outside the zone of the mean intensity $\pm 0.1$[G].

We reduced these 27 feature variables to 13 by applying stepwise-feature selection \cite{Guyon2003} using logistic regression. As a result, 13 feature variables, 1, 2, 6, 7, 9, 11, 12, 13, 15, 20, 21, 24, and 26, were adopted.

For these selected feature vectors, machine learning methods by decision
tree, SVM, and random forest was applied, and in each of these a grid
search were conducted
over the training data to choose the best model.

\subsection{Throughput evaluation}


For evaluating the throughput of the recognition, the time required for the recognition of the entire sequence data was divided by the number of samples of the sequence data to derive the mean value of the recognition throughput per time unit. For the comparative methods, we calculated the computation time of a feature vector in one time window, the time taken to recognize the activity from the feature vector in one time window, and the sums of these values.

%% file: result.tex
\section{Results}
\label{sec:実験}
In the following, we compare the results of evaluating the accuracy of the activity recognition with the DRNN by searching various parameters with those of the accuracy of the existing technique. First, after showing the best model that yielded the highest accuracy, we show the accuracy when the parameters are varied. Furthermore, for the throughput of the recognition, we compare the time required for feature calculation and recognition in the existing technique and the time required until the result is output after inputting one sample in the DRNN.

\subsection{Best model}
\label{sec:The best model}
The best model is the model that showed the best recognition result for the sequence data during the experiment. The parameters selected in the best model are shown in Table \ref{tab:最良モデルパラメータ}.

\begin{figure}[tb]
 \centering
 \includegraphics[width=78mm]{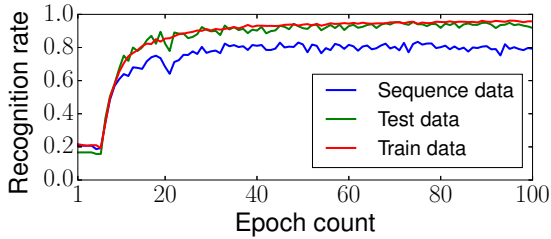}
 \caption{Accuracy transition of the best model}
 \label{fig:最良モデルの正解率推移}
\end{figure}

Fig. \ref{fig:最良モデルの正解率推移} shows the transition of the correct
recognition rate in each epoch. 
The “recognition rate” is derived by the ratio of the correct  
recognition time against total time for each trial, and it is also  
referred to simply as accuracy. 
In the best model, the test recognition rate was $95.42$\% at maximum. The recognition rate for the sequence data was $83.43$\% at maximum.

\begin{table}[b]
\centering
 \fontsize{8pt}{10pt}\selectfont
\caption{Best model parameters}
\label{tab:最良モデルパラメータ}
\begin{tabular}{c||c} \hline
Parameters&Best value\\ \hline \hline

Number of internal layers    &3 \\ \hline
~Number of units in one layer       &60 \\ \hline
Truncated time    &$T=30$      \\ \hline
Gradient clipping parameter  & $c=5$ \\ \hline
Dropout rate&$p=0.5$  \\ \hline
\end{tabular}
\end{table}

In Fig. \ref{fig:最良モデルの正解率推移}, it can be seen that the
recognition rate increases as the epochs proceed. According to the results
in this figure, we judged that it is reasonable to stop the training at
about epoch 80, and for the subsequent evaluations, we extracted the
average recognition rate from epoch 71 to epoch 80 for comparison.

\begin{figure}[b]
 \centering
 \includegraphics[width=75mm]{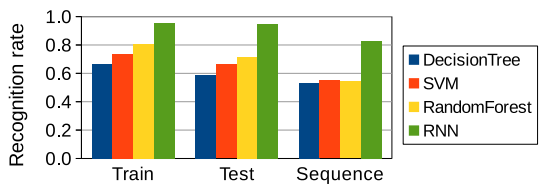}
 \caption{Accuracy of comparative methods and best model}
 \label{fig:推定精度}
\end{figure}

\subsection{Comparison with existing methods}
\label{sec:既存手法との比較}

Fig. \ref{fig:推定精度} shows a graph comparing the accuracy of the existing method with that of the proposed method. The Bar represents the mean recognition rate from epoch 71 to epoch 80. In the test data, the DRNN shows results that are 35.18\%, 27.76\% and 22.35\% better than those of decision tree, SVM, and random forest, respectively. In the sequence data, it shows results that are 28.03\%, 26.04\% and 26.74\% better than those of decision tree, SVM, and random forest, respectively.
\subsection{Varying parameters}
\label{sec:Varying parameters}

\begin{figure}[H]
 \centering
 \includegraphics[width=80mm]{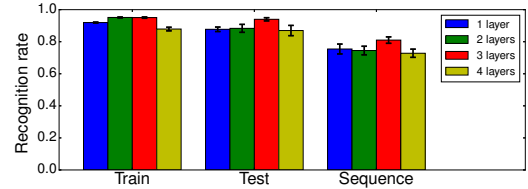}
 \caption{Comparison of accuracy according to the number of internal layers}
 \label{fig:中間層の数による正解率の比較}
\end{figure}

The results obtained by changing the number of internal layers are shown
in Fig. \ref{fig:中間層の数による正解率の比較}. The thin line at the top
of bar represents the standard deviation. We changed only the number of internal layers among the parameters of the best model. As can be seen in the figure, the recognition rate is highest in the case of three layers for any of the training, test, and sequence data. In particular, for the sequence data, the recognition rate is about 8.2\% higher than that of the worst four-layer model.

\begin{figure}[H]
 \centering
 \includegraphics[width=80mm]{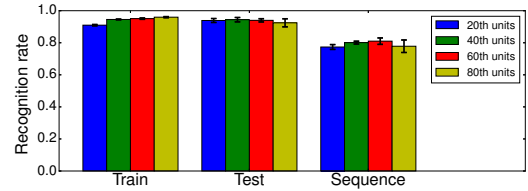}
 \caption{Comparison of accuracy according to the number of units}
 \label{fig:ユニット数による正解率の比較}
\end{figure}

The results obtained by changing the number of internal layer units are shown in Fig. \ref{fig:ユニット数による正解率の比較}. We changed only the number of internal layer units, using 20, 40, 60, and 80 units, among the parameters of the best model. In the figure, it can be seen that the recognition rate is highest in the case of 60 units in test and sequence data. In particular, in the sequence data the recognition rate is about 3.7\% higher than that of the lowest, 20-unit, model.

\begin{figure}[H]
 \centering
 \includegraphics[width=80mm]{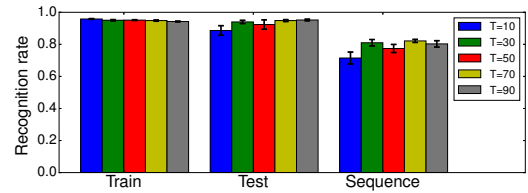}
 \caption{Comparison of accuracy according to the truncated timess}
 \label{fig:truncated timeによる正解率の比較}
\end{figure}

The results of the experiments where the truncated time was changed are shown in Fig. \ref{fig:truncated timeによる正解率の比較}. We changed only the truncated time, using 10, 30, 50, 70, and 90, among the parameters of the best model. The figure shows that the performance is relatively good at $T=30$ or $T=70$, and worst at $T=10$. For the sequence data, a difference of about $10.7$\% occurred between the most accurate $T=70$ model and the $T=10$ model.

\begin{figure}[H]
 \centering
 \includegraphics[width=80mm]{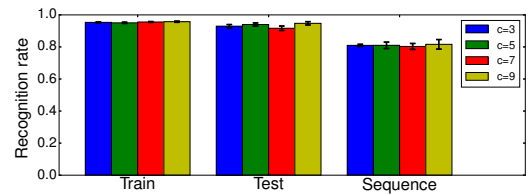}
 \caption{Comparison of accuracy according to the gradient clipping parameters}
 \label{fig:勾配クリッピングパラメータによる正解率の比較}
\end{figure}

\begin{figure}[H]
 \centering
 \includegraphics[width=80mm]{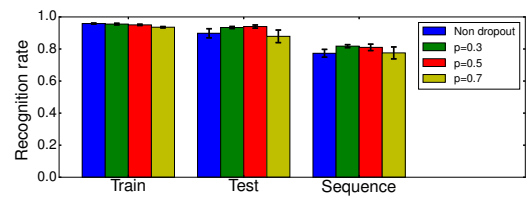}
 \caption{Comparison of accuracy according to dropout rates}
 \label{fig:ドロップアウトによる正解率の比較}
\end{figure}

The results of experiments where the gradient clipping parameter was
changed are shown in Fig. \ref{fig:勾配クリッピングパラメータによる正解率の比較}. We changed only the gradient clipping parameter, using 3, 5,
7, and 9, among the parameters of the best model. The figure shows that
no significant recognition rate difference due to the variation in the gradient
clipping parameter can be observed.

The results of experiments where the dropout probability $p$ was changed are shown in Fig. \ref{fig:ドロップアウトによる正解率の比較}. We changed only the dropout probability $p$, using 0, 0.3, 0.5, and 0.7, among the parameters of the best model. The $p=0.3$ model showed the highest recognition rate for the test and sequence data. In particular, for the sequence data, the recognition rate of this model was about $4.6$\% higher than that of the lowest recognition rate model without dropout.

\begin{table}[b]
\centering
 \fontsize{8pt}{10pt}\selectfont
\caption{Evaluation of throughput}
\label{tab:throughput}
\begin{tabular}{rrrr}
  \hline
 & Feature [ms] & Recognition [ms] & total [ms] \\
\hline
Decision Tree & 11.027 & 0.004 & 11.031 \\
  SVM & 11.027 & 0.123 & 11.150 \\
  Random Forest  & 11.027 & 0.056 & 11.083\\
  RNN & - & 1.347 & 1.347\\
   \hline
\end{tabular}
\end{table}

The above results show that, in the task of activity recognition in this time, a large difference appeared in the recognition rate for five parameters. In particular, for the parameter of truncated time, there is a difference of $10.7$\% between the maximum and minimum recognition rate, revealing that the parameter adjustment is effective.

\subsection{Throughput of activity recognition}
\label {sec:Throughput of activity recognition}
We summarize the throughputs of the proposed method and the existing
method in Table \ref{tab:throughput}. When compared according to only
the calculation time at the time of recognition, the existing method is
$1.342$ [ms] faster, but if compared according to the substantial time,
the proposed method is $9.671$ [ms] faster, because the existing
method requires that the feature vector extracted as a
pre-processing.

\subsection{Result with additional dataset}
\label {sec:他データでの検証}

\begin{figure}[b]
\centering
\includegraphics[width=85mm]{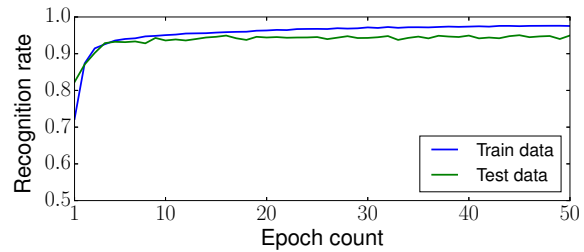}
\caption{Result for HAR dataset}
\label{fig:UCIデータセットに対する実験}
\end{figure}

As a result of applying our method and parameters for HAR dataset, the recognition rate was 95.03\% at 45th epoch.
with the cross validation with first three fourth samples as training
data and last one fourth as test data. Fig. \ref{fig:UCIデータセットに対する実験} shows
the transition of the correct recognition rate in each epoch.

%% file: discussion.tex
\section{Discussion}
\label{sec:Discussion}
We consider the proposed method in the light of the experimental results obtained. In addition, in the following, we give more importance to the accuracy for the test and sequence data than the accuracy for the training data as a basic evaluation criterion.
\subsection{Activity recognition with the best model}
\label{sec:Activity recognition with the best model}
Using the best model, it was possible to perform the recognition with a higher recognition rate and faster response speed than those of the traditional methods. In particular, an recognition rate of $95.42$\% at maximum was obtained for the segmented test data.

\begin{figure*}[htbp]
  \begin{minipage}{0.5\hsize}
    \begin{center}
      \includegraphics[width=96mm]{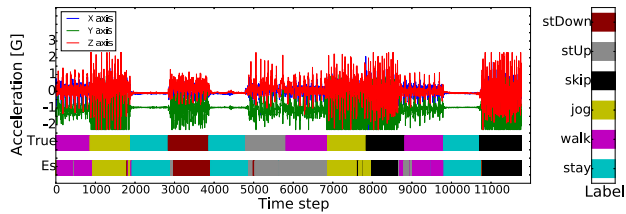}
    (a)~Recognition example 1
\end{center}
  \end{minipage}
  \begin{minipage}{0.5\hsize}
    \begin{center}
      \includegraphics[width=96mm]{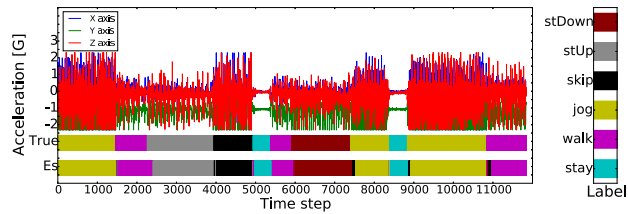}
(b)~Recognition example 2
\end{center}
  \end{minipage}
\hspace{2mm}
 \caption{Visualization of the recognized result}
  \label{fig:推定結果の視覚化}
\end{figure*}

On the other hand, for the sequence data, the recognition rate dropped to $83.43$\% at maximum. The recognized results for the sequence data shown in Fig. \ref{fig:推定結果の視覚化} verify this. The horizontal axis represents 10 [ms] per 1 time in time number and the vertical axis represents acceleration in the gravitational acceleration unit [G]. A color chart displayed as ``True" represents the correct solution label, and a color chart displayed thereunder as ``Es" represents the estimation label. In Fig. \ref{fig:推定結果の視覚化}(a), it can be observed that in general the recognition performance was good, but it caused similar activities to be erroneously recognized at the time $6000-7000$. In addition, it caused a delay in recognition near the time $2300$, as seen  in Fig. \ref{fig:推定結果の視覚化}(b). This phenomenon is considered to have been caused by the fact that unlearned signals that cannot be classified into any activity during the transition of activity were input.


{\color{black}
\subsection{What are the optimal parameters?}
\label{sec:What are the optimal parameters?}
}

\subsubsection{Number of layers and units}
\label{sec:中間層の数と学習の関係の考察}
Because NNs can handle higher order feature vectors, as the
number of layers is increased, the goodness of fit to the training data
is increased. It can be seen in Fig. \ref{fig:中間層の数による正解率の比較} that, in the present experiment, the training accuracy increased as the number of layers was increased up to three in the internal layer. However, the accuracy decreased in the four-layer model. This phenomenon can be interpreted to have occurred as a result of increasing in the learning difficulty by the increase in the excessive number of layers. Further, for the generalization performance, it is known that the accuracy is liable to decrease because of the overfitting when the freedom of the model becomes too high. This fact was also demonstrated in this experiment by the results for the test and sequence data shown in Fig. \ref{fig:中間層の数による正解率の比較}. According to the above, an unnecessary increase in the layers in the model design should be avoided, because it may lead to a reduction in the generalization performance.

Furthermore, when the number of internal layers is increased, the computation time and memory usage are increased. In this experiment, the computation times taken per epoch were $58.89~[{\rm s}]$ with the single layer, $89.59~[{\rm s}]$ with two layers, $116.39~[{\rm s}]$ with three layers, and $144.83~[{\rm s}]$ with 4 layers of the internal layer, respectively. In addition, the throughputs at the time of recognition were $0.512~[{\rm ms}]$ with single layer, $0.909~[{\rm ms}]$ with two layers, $1.347~[{\rm ms}]$ with three layers, and $1.720~[{\rm ms}]$ with four layers of the internal layer, respectively. This time, we chose the best model based on the accuracy, but if we prefer to obtain a high throughput at the expense of accuracy, simplifying the calculation by reducing the layers should be considered.

Almost the same consideration may be also possible for the number of
units as the number of layers. As can be seen in Fig. \ref{fig:ユニット数による正解率の比較}, in order to obtain a high generalization
performance, the number of units should not be excessively
increased. Further, when increasing the number of units, because the
amount of computation time and memory usage increases, an adjustment
will be required when a trade-off is implemented.

\subsubsection{Truncated time}
\label{sec:Truncated timeと学習の関係の考察}
For the truncated time, it can be seen in Fig. \ref{fig:truncated timeによる正解率の比較} that the performance decreased for the test and sequence data at $T=10$, but the optimum value cannot be obtained stably. Here, the recognition rate, which was relatively high at $T=30$ and $T=70$, is about 200 and 470, respectively, if converted into beats per minutes (BPM). It is considered that one cycle of human walking and activities captured by the network may fall within a range of this degree. Thus, in order to capture the features of input signals, it is considered effective to take the variation period of the signals being handled into account when determining the truncated time.

\subsubsection{Gradient clipping parameter}
\label{sec:勾配クリッピングパラメータと学習の関係の考察}
In Fig. \ref{fig:勾配クリッピングパラメータによる正解率の比較}, no significant performance difference generated by the gradient clipping parameters among the data used in this experiment is seen. It is possible that gradient explosion did not occur
to a great extent for the data in this time. In addition, in the DRNN, because the likelihood that gradient explosion will  occur increases as the truncated time and the number of layers increase, it is necessary to suppress the gradient moderately by taking also the values of other parameters into consideration.

\subsubsection{Dropout rate}
\label{sec:ドロップアウト確率と学習の関係の考察}
First, as can be seen in Fig. \ref{fig:ドロップアウトによる正解率の比較} in the cases where the dropout  was and was not applied, applying it
leads to a tendency that the recognition rate is liable to become higher for the
test and sequence data. For the dropout probability, approximately
$50$\%, as frequently used in CNNs, is considered to be appropriate also
in the RNN.

\subsubsection{Validity of parameters}
\label{sec:Validity of parameters}
It is found from Fig.12 that when the parameters adjusted for the recognition of HASC dataset were utilized for the training of the HAR dataset, a high recognition rate of 95.03\% was obtained.
\subsection{Throughput and training time}
\label{throughputと学習時間について}
 The throughput of the recognition was 1.347 [ms], an 8.19 times faster
 speed than that of the existing method. Considering that the data are
 currently acquired at 100 [Hz], this throughput is sufficient to allow
 real time processing. In addition, because the RNN basically performs only the product-sum operation by the number of times of the dimensions of $R$ and $W$ at the time of recognition, it is considered that implementation in low power devices such as smartphones may also be possible in the future.

On the other hand, the training time was $116.39~[{\rm s}]$ per epoch on average. This is a very large and non-negligible amount of time;
however, by calculating the training using a high-speed computer in
advance, high throughput processing may be possible at the time of
recognition. In addition, in principle, because the DRNN can perform the
online learning on a mini-batch basis, by devising a method of feeding
the training data, a high throughput can be expected also in the
training.

The DRNN can achieve both high-speed response with high recognition
rate by using CPU. These advantages are caused by compactly and
small size of the DRNN. In concrete, the total size of the inner
variables and architecture from the input layer to the output layer of
the DRNN is $74166$ pcs, and this is the size of less than 10 \% of the conventional CNN+LSTM model \cite{ordonez2016deep}. Our approach is shown that have a clear advantage for the miniaturization of the devices of recognition. 
\subsection{Future direction}

In this experiment, the activity recognition results of the DRNN were good in terms of recognition rate and throughput. However, new techniques are constantly being developed also for RNNs. Many techniques can be utilized for the activity recognition, such as an approach \cite{LeJH15} using the ReLU function instead of LSTM, a method \cite{Pasa2014} to automatically organize the feature vectors by applying the pre-training used in the CNN, and a method \cite{Wu2015} to accurately estimate the behavior of things by using the physical laws of the real world and its simulation. The verification of these techniques is one of the challenges for future studies.

In addition, in this study the recognition rate of the sequence data
decreased. However, to resolve this issue, it would be possible to apply
the HMM as a post-processing, or to apply a method that takes into
account the context of the label in the RNN, called connectionist
temporal classification \cite{graves2006connectionist}.
And, we develop a compact DRNN circuit to equip into a mobile device.

%% file: conclusion.tex
\section{Conclusion}
\label{sec:Conclusion}
In this paper, the DRNN was constructed for human activity
recognition using raw time series data of acceleration sensors moun-ted
on a mobile device with high recognition rate and high throughput.
The maximum recognition rate was 95.42\% against the test dataset and
was 83.43\% against multiple sequential test dataset. Here, the
maximum recognition rate by traditional methods was 71.65\% and 54.97\%
respectively.
Further, the efficiency of the tuned parameters was confirmed by using the sequential dataset. For the throughput of the recognition per unit time, the constructed DRNN requires only 1.347 [ms], while the traditional method requires 11.031 [ms] which includes 11.027 [ms] for feature extraction. In the future, many techniques, such as a forget-mechanism and pre-training, optimization methodology, and a method that takes the series into consideration, should be studied.



%% file: paper.bbl
\begin{thebibliography}{10}

\bibitem{anguita2013public}
Anguita, D., Ghio, A., Oneto, L., Parra, X., and Reyes-Ortiz, J.~L.
\newblock A public domain dataset for human activity recognition using
  smartphones.
\newblock In {\em ESANN} (2013).

\bibitem{avci2010activity}
Avci, A., Bosch, S., Marin-Perianu, M., Marin-Perianu, R., and Havinga, P.
\newblock {Activity recognition using inertial sensing for healthcare,
  wellbeing and sports applications: A survey}.
\newblock In {\em Architecture of computing systems (ARCS), 2010 23rd
  international conference on}, VDE (2010), 1--10.

\bibitem{Bao2004a}
Bao, L., and Intille, S.~S.
\newblock {\em {Activity Recognition from User-Annotated Acceleration Data}}.
\newblock 2004.

\bibitem{bhattacharya2014using}
Bhattacharya, S., Nurmi, P., Hammerla, N., and Pl{\"{o}}tz, T.
\newblock {Using unlabeled data in a sparse-coding framework for human activity
  recognition}.
\newblock {\em Pervasive and Mobile Computing 15\/} (2014), 242--262.

\bibitem{Bulling2014}
Bulling, A., Blanke, U., and Schiele, B.
\newblock {A tutorial on human activity recognition using body-worn inertial
  sensors}.
\newblock {\em ACM Computing Surveys 46\/} (2014), 1--33.

\bibitem{LSTM99}
Gers, F.~A., Schmidhuber, J., and Cummins, F.
\newblock {Learning to forget: continual prediction with LSTM.}
\newblock {\em Neural computation 12}, 10 (2000), 2451--2471.

\bibitem{graves2006connectionist}
Graves, A., Fern{\'{a}}ndez, S., Gomez, F., and Schmidhuber, J.
\newblock {Connectionist temporal classification: labelling unsegmented
  sequence data with recurrent neural networks}.
\newblock In {\em Proceedings of the 23rd international conference on Machine
  learning}, ACM (2006), 369--376.

\bibitem{Guyon2003}
Guyon, I., and Elisseeff, A.
\newblock {An introduction to variable and feature selection}.
\newblock {\em Journal of Machine Learning Research 3\/} (2003), 1157--1182.

\bibitem{NN}
Haykin, S.
\newblock {Neural networks-A comprehensive foundation}, 1994.

\bibitem{dropout}
Hinton, G.~E., Srivastava, N., Krizhevsky, A., Sutskever, I., and
  Salakhutdinov, R.~R.
\newblock {Improving neural networks by preventing co-adaptation of feature
  detectors}.
\newblock {\em arXiv: 1207.0580\/} (2012), 1--18.

\bibitem{LSTM97}
Hochreiter, S., Hochreiter, S., Schmidhuber, J., and Schmidhuber, J.
\newblock {Long short-term memory.}
\newblock {\em Neural computation 9}, 8 (1997), 1735--80.

\bibitem{ubicomp15sozo}
Inoue, S., Ueda, N., Nohara, Y., and Nakashima, N.
\newblock {Mobile Activity Recognition for a Whole Day: Recognizing Real
  Nursing Activities with Big Dataset}.
\newblock In {\em ACM Int'l Conf. Pervasive and Ubiquitous Computing (Ubicomp)}
  (Osaka, 2015).

\bibitem{Inouye2001}
Inouye, S.~K., Foreman, M.~D., Mion, L.~C., Katz, K.~H., and Cooney, L.~M.
\newblock {Nurses' recognition of delirium and its symptoms: comparison of
  nurse and researcher ratings.}
\newblock {\em Archives of internal medicine 161\/} (2001), 2467--2473.

\bibitem{kawaguchi2011hasc}
Kawaguchi, N., Ogawa, N., Iwasaki, Y., Kaji, K., Terada, T., Murao, K., Inoue,
  S., Kawahara, Y., Sumi, Y., and Nishio, N.
\newblock Hasc challenge: gathering large scale human activity corpus for the
  real-world activity understandings.
\newblock In {\em Proceedings of the 2nd Augmented Human International
  Conference}, ACM (2011), 27.

\bibitem{Kim2010}
Kim, E., Helal, S., and Cook, D.
\newblock {Human Activity Recognition and Pattern Discovery}.
\newblock {\em Pervasive Computing, IEEE 9\/} (2010), 48--53.

\bibitem{adam}
Kingma, D.~P., and Ba, J.~L.
\newblock {Adam: a Method for Stochastic Optimization}.
\newblock {\em International Conference on Learning Representations\/} (2015),
  1--13.

\bibitem{Krishnan2014}
Krishnan, N.~C., and Cook, D.~J.
\newblock {Activity recognition on streaming sensor data}.
\newblock {\em Pervasive and Mobile Computing 10}, PART B (2014), 138--154.

\bibitem{kunze2006towards}
Kunze, K., Barry, M., Heinz, E., Lukowicz, P., Majoe, D., and Gutknecht, J.
\newblock {Towards Recognizing Tai Chi {\{}{\^{A}}{\}}?` An Initial Experiment
  Using Wearable Sensors}.
\newblock In {\em Applied Wearable Computing (IFAWC), 2006 3rd International
  Forum on}, VDE (2006), 1--6.

\bibitem{ladha2013climbax}
Ladha, C., Hammerla, N.~Y., Olivier, P., and Pl{\"{o}}tz, T.
\newblock {ClimbAX: skill assessment for climbing enthusiasts}.
\newblock In {\em Proceedings of the 2013 ACM international joint conference on
  Pervasive and ubiquitous computing}, ACM (2013), 235--244.

\bibitem{Lane2010a}
Lane, N.~D., Miluzzo, E., Lu, H., Peebles, D., Choudhury, T., and Campbell,
  A.~T.
\newblock {A survey of mobile phone sensing}.
\newblock {\em IEEE Communications Magazine 48\/} (2010), 140--150.

\bibitem{LeJH15}
Le, Q.~V., Jaitly, N., and Hinton, G.~E.
\newblock {A Simple Way to Initialize Recurrent Networks of Rectified Linear
  Units}.
\newblock {\em CoRR abs/1504.0\/} (2015).

\bibitem{mazilu2015wearable}
Mazilu, S., Blanke, U., Dorfman, M., Gazit, E., Mirelman, A., {M Hausdorff},
  J., and Tr{\"{o}}ster, G.
\newblock {A Wearable Assistant for Gait Training for Parkinson’s Disease with
  Freezing of Gait in Out-of-the-Lab Environments}.
\newblock {\em ACM Transactions on Interactive Intelligent Systems (TiiS) 5}, 1
  (2015), 5.

\bibitem{ordonez2016deep}
Ord{\'o}{\~n}ez, F.~J., and Roggen, D.
\newblock Deep convolutional and lstm recurrent neural networks for multimodal
  wearable activity recognition.
\newblock {\em Sensors 16}, 1 (2016), 115.

\bibitem{Pasa2014}
Pasa, L., and Sperduti, A.
\newblock {Pre-training of Recurrent Neural Networks via Linear Autoencoders}.
\newblock {\em Advances in Neural Information Processing Systems 27\/} (2014),
  3572--3580.

\bibitem{clipping}
Pascanu, R., Mikolov, T., and Bengio, Y.
\newblock {On the difficulty of training recurrent neural networks}.
\newblock {\em Proceedings of The 30th International Conference on Machine
  Learning}, 2 (2012), 1310--1318.

\bibitem{Saeedi2014}
Saeedi, R., Schimert, B., and Ghasemzadeh, H.
\newblock {Cost-Sensitive Feature Selection for On-Body Sensor Localization}.
\newblock {\em 2nd International Workshop on Human Activity Sensing Corpus and
  its Application (HASCA2014) held at UbiComp 2014\/} (2014), 833--842.

\bibitem{Strohrmann2011}
Strohrmann, C., Harms, H., and Tr{\"{o}}ster, G.
\newblock {What do sensors know about your running performance?}
\newblock In {\em Proceedings - International Symposium on Wearable Computers,
  ISWC} (2011), 101--104.

\bibitem{T-BPTT}
Sutskever, I.
\newblock {Training Recurrent neural Networks}.
\newblock {\em PhD thesis\/} (2013), 101.

\bibitem{tokui2015chainer}
Tokui, S., Oono, K., Hido, S., and Clayton, J.
\newblock Chainer: a next-generation open source framework for deep learning.
\newblock In {\em Proceedings of Workshop on Machine Learning Systems
  (LearningSys) in The Twenty-ninth Annual Conference on Neural Information
  Processing Systems (NIPS)} (2015).

\bibitem{Wu2015}
Wu, J., Yildirim, I., Lim, J., Freeman, W., and Tenenbaum, J.
\newblock {Galileo : Perceiving Physical Object Properties by Integrating a
  Physics Engine with Deep Learning}.
\newblock {\em Advances in Neural Information Processing Systems 28 (NIPS
  2015)\/} (2015), 1--9.

\bibitem{yang2015deep}
Yang, J.~B., Nguyen, M.~N., San, P.~P., Li, X.~L., and Krishnaswamy, S.
\newblock Deep convolutional neural networks on multichannel time series for
  human activity recognition.
\newblock In {\em Proceedings of the 24th International Joint Conference on
  Artificial Intelligence (IJCAI), Buenos Aires, Argentina} (2015), 25--31.

\bibitem{Zhan2014}
Zhan, K., Faux, S., and Ramos, F.
\newblock {Multi-scale Conditional Random Fields for first-person activity
  recognition}.
\newblock {\em Pervasive Computing and {\ldots}\/} (2014).

\bibitem{Zhang2012}
Zhang, M., and Sawchuk, A.
\newblock {Motion primitive-based human activity recognition using a
  bag-of-features approach}.
\newblock {\em Proceedings of the 2nd ACM SIGHIT {\ldots}}, 1 (2012), 631.

\bibitem{Zhang2011}
Zhang, M., and Sawchuk, A.~A.
\newblock {A feature selection-based framework for human activity recognition
  using wearable multimodal sensors}.
\newblock In {\em Int. Conf. Body Area Networks} (2011), 92--98.

\end{thebibliography}
